\documentclass[letterpaper]{article} 
\usepackage{aaai23}  
\usepackage{times}  
\usepackage{helvet}  
\usepackage{courier}  
\usepackage[hyphens]{url}  
\usepackage{graphicx} 
\usepackage{natbib}  
\usepackage{caption} 
\usepackage{algorithm}
\usepackage{algorithmic}
\usepackage{amssymb}
\usepackage{multirow}
\usepackage{newfloat}
\usepackage{listings}
\DeclareCaptionStyle{ruled}{labelfont=normalfont,labelsep=colon,strut=off} 
\floatstyle{ruled}
\newfloat{listing}{tb}{lst}{}
\floatname{listing}{Listing}
\title{SwinRDM: Integrate SwinRNN with Diffusion Model towards High-Resolution and High-Quality Weather Forecasting}
\author{
    Lei Chen\equalcontrib,
    Fei Du\equalcontrib\thanks{Corresponding author.},
    Yuan Hu,
    Zhibin Wang,
    Fan Wang
}
\affiliations{
    Alibaba Group \\

    \{fanjiang.cl, dufei.df, lavender.hy, zhibin.waz, fan.w\}@alibaba-inc.com
}

\usepackage{bibentry}

\begin{document}

\maketitle

\begin{abstract}
Data-driven medium-range weather forecasting has attracted much attention in recent years. However, the forecasting accuracy at high resolution is unsatisfactory currently. Pursuing high-resolution and high-quality weather forecasting, we develop a data-driven model SwinRDM which integrates an improved version of SwinRNN with a diffusion model. SwinRDM performs predictions at 0.25-degree resolution and achieves superior forecasting accuracy to IFS (Integrated Forecast System), the state-of-the-art operational NWP model, on representative atmospheric variables including 500 hPa geopotential (Z500), 850 hPa temperature (T850), 2-m temperature (T2M), and total precipitation (TP), at lead times of up to 5 days. We propose to leverage a two-step strategy to achieve high-resolution predictions at 0.25-degree considering the trade-off between computation memory and forecasting accuracy. Recurrent predictions for future atmospheric fields are firstly performed at 1.40625-degree resolution, and then a diffusion-based super-resolution model is leveraged to recover the high spatial resolution and finer-scale atmospheric details. SwinRDM pushes forward the performance and potential of data-driven models for a large margin towards operational applications.
\end{abstract}


\section{Introduction}

Accurate weather forecasting is beneficial to human beings in several areas such as agriculture, energy, and public transportation. Numerical Weather Prediction (NWP) has long been adopted for weather forecasting. It has been improved  by better physics parameterization techniques and high-quality atmospheric observations in the past few decades. However, this approach requires huge amounts of computing power, which may limit its application in industry.

With the development of machine learning (ML), especially deep learning (DL) techniques, many studies are employing data-driven DL methods to forecast atmospheric variables. The purely data-driven DL models are often orders of magnitude faster than the NWP model. However, the performance of current DL models is unsatisfactory currently. To facilitate the development of data-driven weather forecasting, some benchmarks \cite{rasp2020weatherbench,garg2022weatherbench,de2020rainbench} are constructed to enable a thorough comparison of different methods. Among them, WeatherBench \cite{rasp2020weatherbench} is one of the widely used benchmarks, which is constructed by regridding the ERA5 reanalysis dataset \cite{hersbach2020era5} from $0.25^{\circ}$ resolution to three different resolutions (i.e., $5.625^{\circ}$, $2.8125^{\circ}$ and $1.40625^{\circ}$). It focuses on the medium-range global prediction of a few key variables at lead times of up to 5 days.
Several works have tried to improve the prediction performance on WeatherBench \cite{rasp2021data, weyn2020improving, hu2022swinvrnn}.
Among them, SwinVRNN \cite{hu2022swinvrnn} achieves the best performance by integrating a variational recurrent neural network (SwinRNN) with a feature perturbation module.


Although these works have achieved great success in global weather forecasting, their methods are built on low-resolution (usually lower than $1^{\circ}$) data. The largest resolution of WeatherBench is $1.40625^{\circ}$, corresponding to a $128 \times 256$ pixels grid. And the distance between every two pixels is larger than 100km, which is too coarse for a forecasting model to capture the fine-scale dynamics \cite{pathak2022fourcastnet}. \cite{keisler2022forecasting} builds a graph neural network (GNN) to forecast global weather on the 1-degree scale. It shows comparable performance on wind and relative humidity to the IFS. However, its resolution is still relatively low. FourCastNet \cite{pathak2022fourcastnet} trains an Adaptive Fourier Neural Operator (AFNO) model directly at $0.25^{\circ}$ resolution ERA5 dataset, which achieves comparable performance to the IFS at short-range lead times and can resolve many important small-scale phenomena. However, there still exists a performance gap between data-driven models and the IFS at lead times of up to 5 days, especially on representative variables such as Z500 and T850.

In this paper, we focus on building a global weather forecasting model at $0.25^{\circ}$ resolution and propose a SwinRDM model by integrating an improved SwinRNN \cite{hu2022swinvrnn} with a diffusion-based super-resolution model. Since SwinRNN achieves superior performance at low resolution, we employ it as our base model. We experimentally analyze the SwinRNN model and build an improved version named SwinRNN+ by replacing the multi-scale network with a single-scale design and adding a feature aggregation layer. Our SwinRNN+ achieves higher performance than IFS on all key variables at lead times of up to 5 days at $1.40625^{\circ}$ resolution. Note that this is a considerable improvement compared to SwinRNN that can only compete with the IFS model on surface-level variables at $5.625^{\circ}$ resolution. Different from FourCastNet, to generate high-resolution global weather prediction, we resort to the super-resolution (SR) technique rather than directly train the SwinRNN+ model on $0.25^{\circ}$ resolution data due to the prohibitive computational cost. The super-resolution task is implemented using a conditional diffusion model \cite{saharia2021image}, which trains a U-Net model \cite{ronneberger2015u} to iteratively refine the outputs starting from pure Gaussian noises. This model is shown to be able to generate photo-realistic outputs compared to traditional super-resolution models \cite{saharia2021image}. We show in this work that the diffusion model-based super-resolution conditioned on low-resolution predictions can capture small-scale variations and generate high-quality weather forecasting at high resolution.

Our contribution can be summarized as follows:

\begin{itemize}
\item We conduct experimental studies on the SwinRNN model and propose an improved version --- SwinRNN+. It achieves superior performance than the state-of-the-art IFS model on all representative variables at the resolution of $1.40625^{\circ}$ and lead times of up to 5 days.
\item We employ a conditional diffusion model for super-resolution conditioned on SwinRNN+ outputs, which achieves high-quality weather forecasting at the resolution of $0.25^{\circ}$ with an optimal trade-off between computation cost and forecast accuracy.
\item Experimental results on the ERA5 dataset show that our SwinRDM model not only outperforms IFS but also achieves high-quality forecasts with finer-scale details, which sets a solid baseline for data-driven DL weather forecasting models.
\end{itemize}

\section{Related Works}
In this section, we briefly review some deep learning-based weather forecasting methods and super-resolution methods.
\subsection{Deep Learning-based Weather Forecasting}
Deep learning has been investigated to perform data-driven weather forecasting in recent years, and the goal is to fully replace the NWP model. 
Some works focus on a particular local area \cite{shi2017deep,sonderby2020metnet,bihlo2021generative}. However, using data in a local spatial domain may result in uncertainty around the boundary regions \cite{bihlo2021generative}. For global weather forecasting, some widely used networks in computer vision has been applied, including ResNet \cite{rasp2021data}, U-Net \cite{weyn2020improving}, VAE \cite{hu2022swinvrnn}, and GNNs \cite{keisler2022forecasting}. Among global weather forecasting methods, SwinVRNN \cite{hu2022swinvrnn} is the first work that can compete with the IFS on representative surface-level variables (T2M and TP) at lead times of up to 5 days. It constructs a deterministic SwinRNN model based on the Swin Transformer block \cite{liu2021swin} and builds a perturbation module to perform ensemble forecasting. However, the resolution of SwinVRNN is relatively low, and it cannot achieve comparable performance on pressure-level variables such as Z500 and T850. FourCastNet \cite{pathak2022fourcastnet} is the first work that directly builds the network on the $0.25^{\circ}$ resolution data. Although it achieves high performance on short timescales, it cannot compete with the IFS at a 5-day lead time. In this paper, we also intend to predict the atmospheric variables at $0.25^{\circ}$ resolution. We improve the SwinRNN network at low resolution and employ the diffusion-based super-resolution model to generate high-resolution and high-quality results.

\begin{figure*}[ht]
    \centering
    \includegraphics[width=\linewidth]{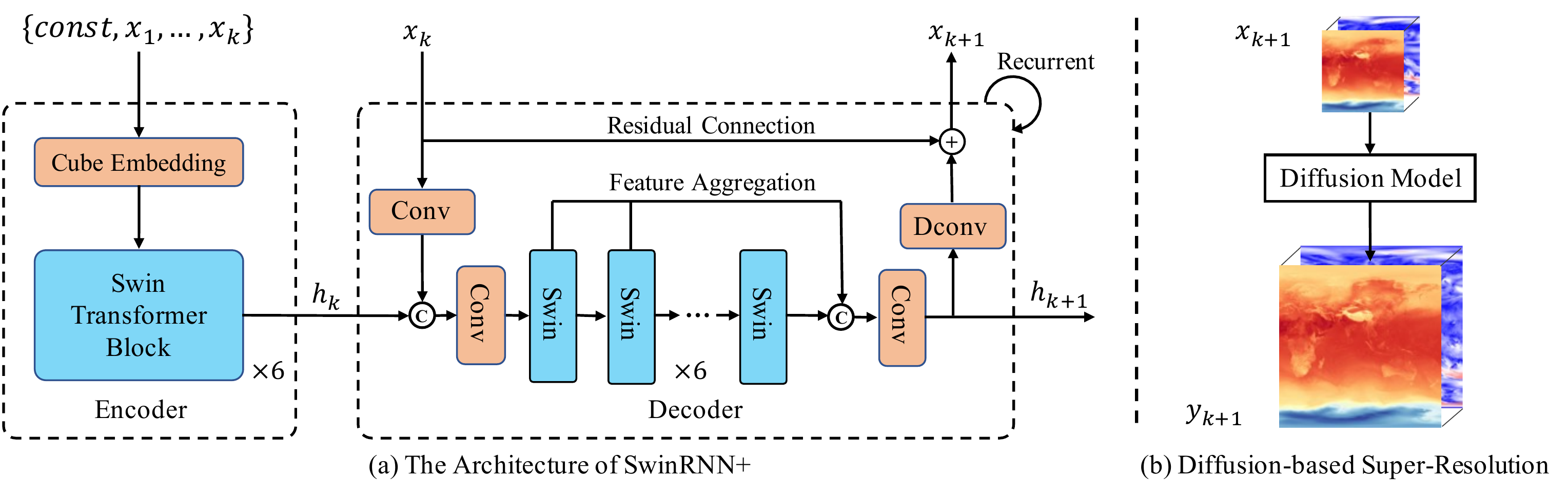}
    \caption{The SwinRDM consists of two parts: (a) the low-resolution forecasting model SwinRNN+ is an improved version of SwinRNN, which adopts a single-scale architecture and adds a multi-layer feature aggregation component, and (b) the diffusion-based super-resolution model conditions on the prediction $x_{k+1}$ from SwinRNN+.}
    \label{fig:swinrdm}
\end{figure*}

\subsection{Super Resolution}
The goal of super-resolution is to reconstruct a high-resolution image from a low-resolution counterpart. The simplest way to realize super-resolution is interpolation. It is computationally efficient but often suffers from detail loss in regions with complex textures \cite{chen2022arm}. SRCNN \cite{dong2015image} is a pioneering work that exploits CNNs to perform super-resolution. Later, many deep learning methods \cite{kim2016accurate,lim2017enhanced,zhang2018image,liang2021swinir} have been proposed to improve super-resolution performance. These methods usually employ a reconstruction loss such as MSE loss to train the network. However, the network simply trained with a reconstruction loss can hardly capture high texture details and generate perceptually satisfactory results \cite{bihlo2021generative}. Some works \cite{ledig2017photo,wang2018esrgan,wang2021real} employ generative adversarial networks (GANs) \cite{goodfellow2014generative} to encourage the generator to output high-resolution images that are hard to distinguish from the real high-resolution images. Although these methods can generate high-quality images, GAN-based methods are difficult to optimize \cite{arjovsky2017wasserstein}. 

Recently, diffusion models \cite{sohl2015deep} have attracted much attention in image generation since it is able to generate high-quality images \cite{song2019generative, ho2020denoising, song2020improved} that are comparable to GANs. A U-Net architecture is trained with a denoising objective to iteratively refine the outputs starting from pure Gaussian noise. Diffusion models have also been successfully applied to image super-resolution, such as SR3 \cite{saharia2021image} which employs a diffusion model to generate realistic high-resolution images conditioned on low-resolution images. Since the diffusion model is proven to generate high-quality images, we exploit this model in weather forecasting to generate high-quality and high-resolution predictions.

\section{Methodology}
We formulate the high-resolution weather forecasting problem as a combination of weather forecasting at low resolution and super-resolution to high resolution. The proposed model first recurrently forecasts the future atmospheric variables via a recurrent neural network and then reconstructs the high-resolution results from the predicted low-resolution counterparts via a super-resolution network. The recurrent neural network is built based on SwinRNN \cite{hu2022swinvrnn}. We experimentally analyze the architecture of SwinRNN and design an improved version named SwinRNN+ which considerably improves the performance of SwinRNN at $1.40625^{\circ}$ resolution. However, SwinRNN+ can hardly be directly trained at $0.25^{\circ}$ because the training recurrent steps have to be reduced due to limited computational resources, which will lead to inferior performance. Thus, we adopt the super-resolution technique to achieve high-resolution prediction. To generate high-resolution and high-quality results at $0.25^{\circ}$ resolution, we build our super-resolution model based on the diffusion model, which can help capture fine-grained scales compared to traditional super-resolution methods. The architecture of our method is demonstrated in Figure \ref{fig:swinrdm}.

\subsection{Background on SwinRNN} The SwinRNN mainly consists of a multi-scale encoder for historical context information extraction and a multi-scale decoder for hidden state propagation and variable prediction at each recurrent step. The atmospheric variables at each time step are stacked together with a shape $C_{in} \times H \times W$, where $C_{in}$ denotes the number of atmospheric variables, and $H \times W$ denotes the global grid size. 
The stacked result can be regarded as a multi-channel frame similar to many computer vision tasks. 
The encoder takes $k$ consecutive historical frames as input. It first employs a 3D convolutional network-based cube embedding block to project all frames to features with a size $C \times H \times W$ and then extracts four-scale features $(h_k^1, h_k^2, h_k^3, h_k^4)$ via a hierarchical Swin Transformer. The historical context information is embedded in the features, and they are used to initialize the hidden states of the decoder at time step $k$. Then in each future time step, the decoder takes as input the combination of hidden states $(h_k^1, h_k^2, h_k^3, h_k^4)$ and current frame $x_k$, and outputs the updated hidden states $(h_{k+1}^1, h_{k+1}^2, h_{k+1}^3, h_{k+1}^4)$ for next time step and the predicted frame $x_{k+1}$. 
The results of SwinRNN show that it achieves higher performance than the IFS on T2M and TP variables.

Although SwinRNN achieves high performance on surface-level variables, it is only trained at $5.625^{\circ}$ (32 $\times$ 64) resolution and cannot compete with the IFS model on pressure-level variables such as Z500 and T850. Based on its Swin Transformer-based recurrent structure, we propose SwinRNN+ that is able to achieve superior performance on all key surface-level and pressure-level variables than the IFS at $1.40625^{\circ}$ resolution.

\subsection{SwinRNN+}
It is non-trivial to transfer the SwinRNN to high-resolution data even at $1.40625^{\circ}$ (128 $\times$ 256) since the memory consumption in the training stage is quadratic to resolution (e.g., 1 vs. 16 for $5.625^{\circ}$ vs. $1.40625^{\circ}$) for the Swin Transformer architecture. Thus, it is important to balance the capacity of the network and the computational cost. The improvements of our SwinRNN+ over SwinRNN are two folds. First, we replace the multi-scale network with a single-scale network with higher feature dimensions. Second, we fuse the output features of all Swin Transformer layers in the decoder to generate the hidden states and the output predictions.

\subsubsection{Trade Multi-Scale Design for Higher Feature Dimensions.} To increase the capacity of the network, a straightforward way is to increase the dimension of the feature. However, the memory cost increases dramatically with the increase of the feature dimension since there are several recurrent steps during training. We observed from \cite{hu2022swinvrnn} that SwinRNN benefits little from the multi-scale architecture, whereas the structure significantly increases the number of parameters and memory consumption. Thus, we conduct an ablation experiment to compare the performance of multi-scale and single-scale structures on different feature dimensions. The experimental results show that the multi-scale network generally achieves better performance compared to the single-scale network with the same feature dimensions. However, with the increase of the feature dimension, the performance gap between the two different structures is narrowed rapidly, whereas the memory and the parameters of the multi-scale structure increase dramatically. Notably, the single-scale network with a high feature dimension shows better performance and higher efficiency compared to the multi-scale network with a low feature dimension. Thus, we draw a conclusion that the multi-scale architecture limits the potential of SwinRNN, and it is more effective to increase the feature dimension of the single-scale network rather than use a multi-scale structure.

\subsubsection{Multi-Layer Feature Aggregation.} Our second improvement is to aggregate features from multiple layers to learn the hidden states. As shown in Figure \ref{fig:swinrdm}, the decoder fuses features from the 6 layers to update the hidden states via a convolutional layer, while the original SwinRNN only treats the features from the final layer as the hidden states. Our multi-layer feature aggregation network has two advantages compared to SwinRNN. First, the representation power of the hidden states is improved, which is beneficial to the prediction in the current time step and feature propagation to the next time step. Second, the gradient backward propagation path is reduced, and the information can be more easily propagated back to former time steps, which eases the optimization of the recurrent network.

As shown in Figure \ref{fig:swinrdm}, the proposed network consists of 6 Swin Transformer blocks with the same scale in both the encoder and decoder. To enable training on $1.40625^{\circ}$ resolution data, we first use a patch size of $2 \times 2$ to split the image into non-overlapping patches. This is achieved by a convolutional layer with a kernel size of 2 and a stride of 2 in the cube embedding block. Thus, we have a hidden state $h_k$ with a size of $C \times H/2 \times W/2$. In the decoder, the frame $x_k$ is also embedded with a convolutional layer with the same settings, and $x_{k+1}$ is predicted by a transposed convolutional layer. Features from all layers in the decoder are aggregated to increase the representation power. 

\subsection{Diffusion Model for Super Resolution}
In order to achieve high-resolution weather forecasting, we integrate the SwinRNN+ with a super-resolution component based on the diffusion model since it is able to generate realistic images with rich details \cite{saharia2021image}, which can help resolve fine-grained features and generate high-quality and high-resolution forecasting results. 

Diffusion models \cite{ho2020denoising} are a class of generative models consisting of a forward process (or diffusion process) that destroys the training data by successive addition of Gaussian noise and a reverse process that learns to recover the data by reversing this noising process. More specifically, given a sample from data distribution $y_0 \sim q(y_0)$, the diffusion process is a Markov chain that gradually adds Gaussian noise to the data according to a fixed variance schedule $\beta_1, \cdots, \beta_T$:
\begin{equation} \label{diffusion}
    q(y_t | y_{t-1}) = \mathcal{N}(y_t; \sqrt{1 - \beta_{t}}y_{t-1}, (\beta_t)\mathcal{I}).
\end{equation}
If the magnitude $\beta_t$ of the noise added at each step is small enough, and the total step $T$ is large enough, then $y_T$ is equivalent to an isotropic Gaussian distribution. It is convenient to produce samples from a Gaussian noise input $y_T \sim \mathcal{N}(0, \mathcal{I})$ by reversing the above forward process. However, the posterior $q(y_{t-1}|y_t)$ need for sampling is hard to compute, and we need to learn a model parameterized by $\theta$ to approximate these conditional probabilities: 
\begin{equation} \label{reverse}
    p_\theta(y_{t-1} | y_T) =\mathcal{N}(\mu_{\theta}(y_t), \Sigma_\theta(y_t)).
\end{equation}
While there exists a tractable variational lower-bound on $logp_\theta(y_0)$, better results arise from optimizing a surrogate denoising objective:
\begin{equation} \label{e-prediction}
    E_{\epsilon \sim N(0, I), t \sim [0, T]}[\|\epsilon - \epsilon_\theta(y_t, t)\|^2],
\end{equation}
where $y_t \sim q(y_t | y_0)$ is obtained by applying Gaussian noise $\epsilon$ to $y_0$, and $\epsilon_\theta$ is the model to predict the added noise. 

Diffusion models can be conditioned on class labels, text, or low-resolution images \cite{dhariwal2021diffusion, ho2022cascaded, saharia2021image, saharia2021palette, whang2022deblurring, nichol2021glide, ramesh2022hierarchical}. In our case, we make it condition on the low-resolution output of SwinRNN+ for the super-resolution task. During training, the low-resolution data $x_k$ is generated on the fly, and forecasting quality is decreased with the lead time. 
To account for such variation, the model additionally conditions on the time step $k$ of SwinRNN+. Thus, we have a posterior
\begin{equation} \label{posterior}
    p_\theta(y_k^{(t-1)}| y_k^{(t)}, x_{k}, t, k), 
\end{equation}
where $y_k$ is the corresponding high-resolution target of $x_k$, Instead of using the $\epsilon$-prediction formulation, we predict the original targets $y_k$ directly, following \cite{ramesh2022hierarchical}. The model acts like a denoising function, trained using a mean squared error loss:
\begin{equation} \label{x-prediction}
    E_{t \sim [0, T], y_k^{(t)} \sim q_t}[\|y_{k} - f_\theta(y_k^{(t)}, x_{k}, t, k)\|^2].
\end{equation}

For the sampling process, given the same low-resolution input $x_k$, the diffusion model can produce diverse outputs $y_k$ starting from different Gaussian noise samples. Such property makes it possible to perform ensemble forecasting, which is an effective way to improve forecast skills. Thus, we achieve super-resolution and ensemble forecasting at the same time with the diffusion model.

\begin{table}[t]
 \centering
 \tabcolsep=2pt
 \begin{tabular}{cccccc}
 \hline
   Dim. & Fusion & Z500($m^2s^{-2}$) & T850($T$) & T2M($T$) & TP($mm$) \\
\hline
128 &  & 456 & 2.354	& 2.196 & 2.265 \\
128 & $\checkmark$ & 386 & 2.050	& 1.926 & 2.182 \\
256 &  & 394 & 2.092	& 1.957 & 2.207 \\
256 & $\checkmark$ & 371 & 1.971	& 1.843 & 2.148 \\
\hline
 \end{tabular}
 \caption{The RMSE results of the SwinRNN+ with and without multi-layer feature aggregation on different feature dimensions.}
 \label{featfusion}
 \end{table}

\begin{table}[t]
 \centering
 \tabcolsep=4pt
 \begin{tabular}{c  c  c  c  c  c  c }
 \hline
   Feat. Dim. & Multi-Scale & Mem. & Params & Z500 & T850 \\
\hline
    128	& &	12.0G &	4.0M	&417&	2.182\\
128	& $\checkmark$ &13.1G&	62.5M	&	386	&2.050\\
256	& &	14.1G	&15.5M&374&	1.998 \\
256	& $\checkmark$ & 23.7G	&248.6M	&371&	1.971\\
384	& &	19.4G&	34.8M	&	359&	1.929\\
512	& &	25.2G&	61.6M	& 354&	1.912\\
\hline
 \end{tabular}
 \caption{Comparison (RMSE) between the single-scale and multi-scale architectures. The units are the same as Table \ref{featfusion}.}
 \label{multiscale}
 \end{table}

\section{Experiments}

\subsection{Experimental Setup}
\subsubsection{Dataset}
We evaluate our proposed SwinRDM method on the ERA5 dataset \cite{hersbach2020era5} provided by the ECMWF. ERA5 dataset is an atmospheric reanalysis dataset, which consists of atmospheric variables at a $0.25^\circ$ spatial resolution from 1979 to the present day. Data from 1979 to 2016 are chosen as the training set, and data from 2017 and 2018 are used for evaluation, following \cite{rasp2020weatherbench}. We sub-sample the dataset at 6-hour intervals to train our model as in \cite{pathak2022fourcastnet}. The input signal of the SwinRDM contains 71 variables, including geopotential, temperature, relative humidity, longitude-direction wind, and latitude-direction wind at 13 vertical layers, four single-layer fields (2m temperature, 10m wind, and total precipitation), and two constant fields (land-sea mask and orography). 
\subsubsection{Evaluation Metrics}
We follow \cite{rasp2020weatherbench} to evaluate forecast quality using latitude-weighted RMSE (root-mean-square error). In addition, we adopt Fr\'echet inception distance (FID) \cite{heusel2017gans} to quantitatively assess the sample fidelity of the super-resolution outputs.
\subsubsection{Implementation Details} We end-to-end train two models (i.e., SwinRNN+ and SwinRDM) for 50 epochs with a batch size of 16. SwinRNN+ is trained on $1.40625^\circ$ data, while SwinRDM takes $1.40625^\circ$ data as input and outputs $0.25^\circ$ predictions. During training, our models take 6 historical frames as input and recurrently predict 20 frames at 6-hour intervals. For SwinRDM, we randomly select one of them to train the diffusion-based super-resolution model. The dimensions of the feature for the encoder and the decoder are set to 768 and 512, respectively. The cosine learning rate policy is used with initial learning rates of 0.0003 for SwinRNN+ and 0.0002 for the diffusion model. The models are optimized by AdamW using PyTorch on 8 NVIDIA A100 GPUs. For the diffusion model, we adopt the implementation in \cite{nichol2021improved} and use 10 sampling steps to get decent results during inference.

 \begin{table}[t]
 \centering
 \tabcolsep=6pt
 \begin{tabular}{l  c  c  c  c  c  c c c c c c c}
 \hline
 Methods  & FID  & Z500 & T850 & T2M & TP  \\
\hline
    Bilinear  & 235 & 374 & 2.00 & 1.89 & 2.18 \\
    SwinIR  & 227 & 376 & 1.99 & 1.68 & 2.16 \\
    SwinRDM  & 60 & 378 & 2.11 & 1.73 & 2.79 \\
    SwinRDM*  & 145 & 374 & 2.02 & 1.65 & 2.30 \\    
\hline
 \end{tabular}
 \caption{Quantitative comparison (FID and RMSE) of different methods at a lead time of 120 hours. The units are the same as Table \ref{featfusion}.}
 
 \label{table:superres}
 \end{table}
 

 \begin{figure}[t]
    \centering
    \small
    \includegraphics[width=0.95\linewidth]{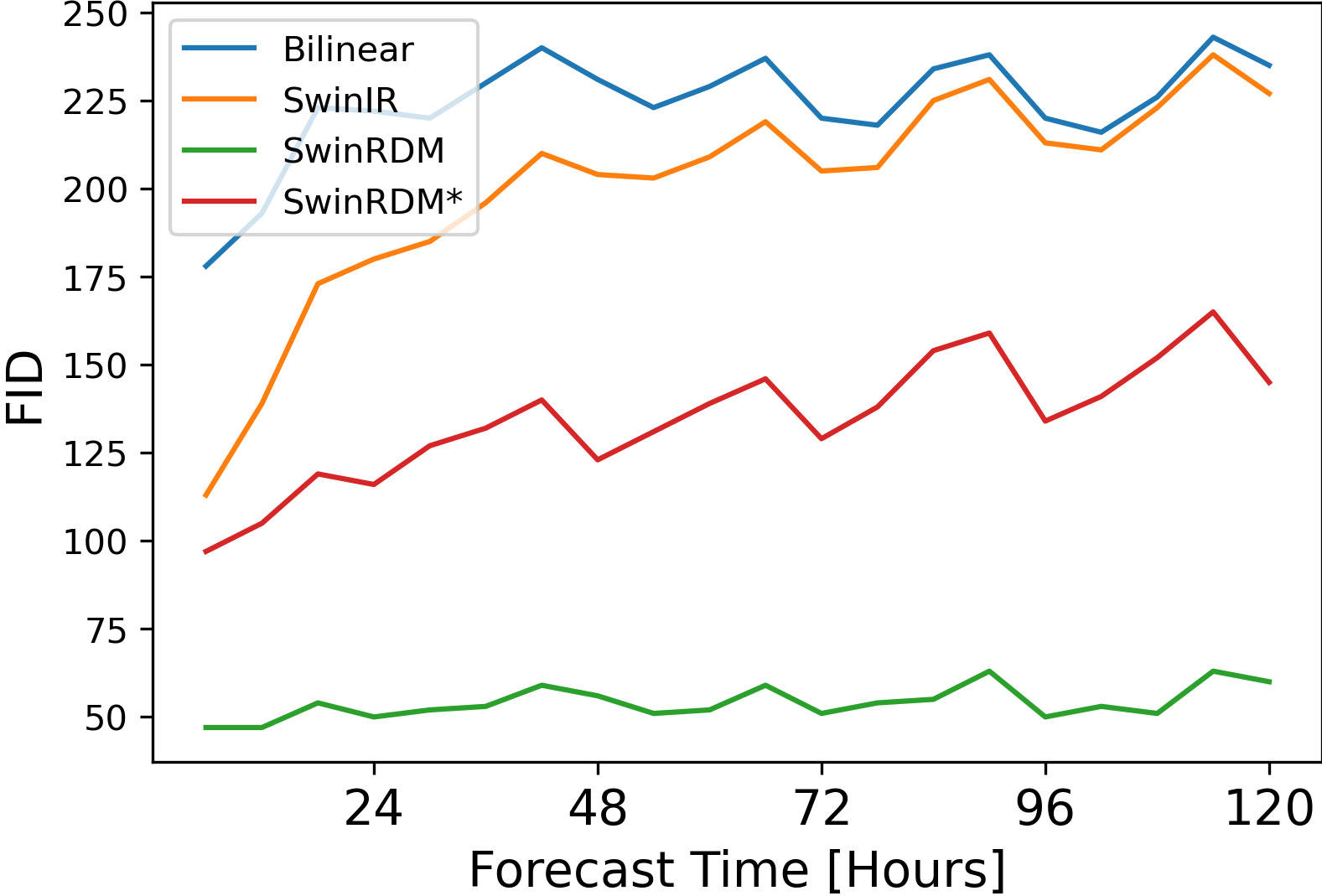}
    \caption{FID in different forecasting times.}
    \label{fig:fid}
\end{figure}

\subsection{Ablation Study on SwinRNN+}
For efficient ablation studies, we train all networks for only 10 epochs.
\subsubsection{Effectiveness of multi-layer feature aggregation}
The multi-layer feature aggregation component fuses features from all Swin Transformer layers of the decoder to generate the hidden states, which can help improve the representation power and ease the optimization process. Table \ref{featfusion} shows the comparison of SwinRNN+ with and without the aggregation layer. The aggregation layer improves the performance of the networks with different feature dimensions, which verifies the effectiveness of the multi-layer feature aggregation.

\subsubsection{Trade-off between multi-scale design and higher feature dimensions}
The number of parameters and the training memory cost increases significantly with the feature dimensions for the multi-scale architecture of the original SwinRNN, which limits the improvement of the model capacity since the computational cost is prohibitive if we use a large feature dimension. We conduct ablation experiments to compare the performance and potential of the multi-scale structure and the single-scale structure. Table \ref{multiscale} shows the comparison results. For both the encoder and decoder, the multi-scale network contains 4 scales, and each scale consists of 6 Swin Transformer blocks, while the single-scale network only has one scale with 6 Swin Transformer blocks. As can be seen from the table, when the feature dimension is set to 128, there exists a large performance gap between the multi-scale structure and the single-scale structure. However, the performance gap is narrowed significantly when the feature dimension increases from 128 to 256. Meanwhile, the training memory and the parameters of the multi-scale structure increase dramatically with the feature dimensions. It is noticeable that the single-scale network with 384-dimensional features shows better performance and less memory cost compared to the multi-scale network with 256-dimensional features. The dimensions can be set to 512 for the single-scale network, while it is unaffordable for the multi-scale one. Thus, instead of using a multi-scale structure to improve the model capacity, it is more effective to increase the feature dimensions.

\begin{figure}[t]
    \centering
    \includegraphics[width=\linewidth]{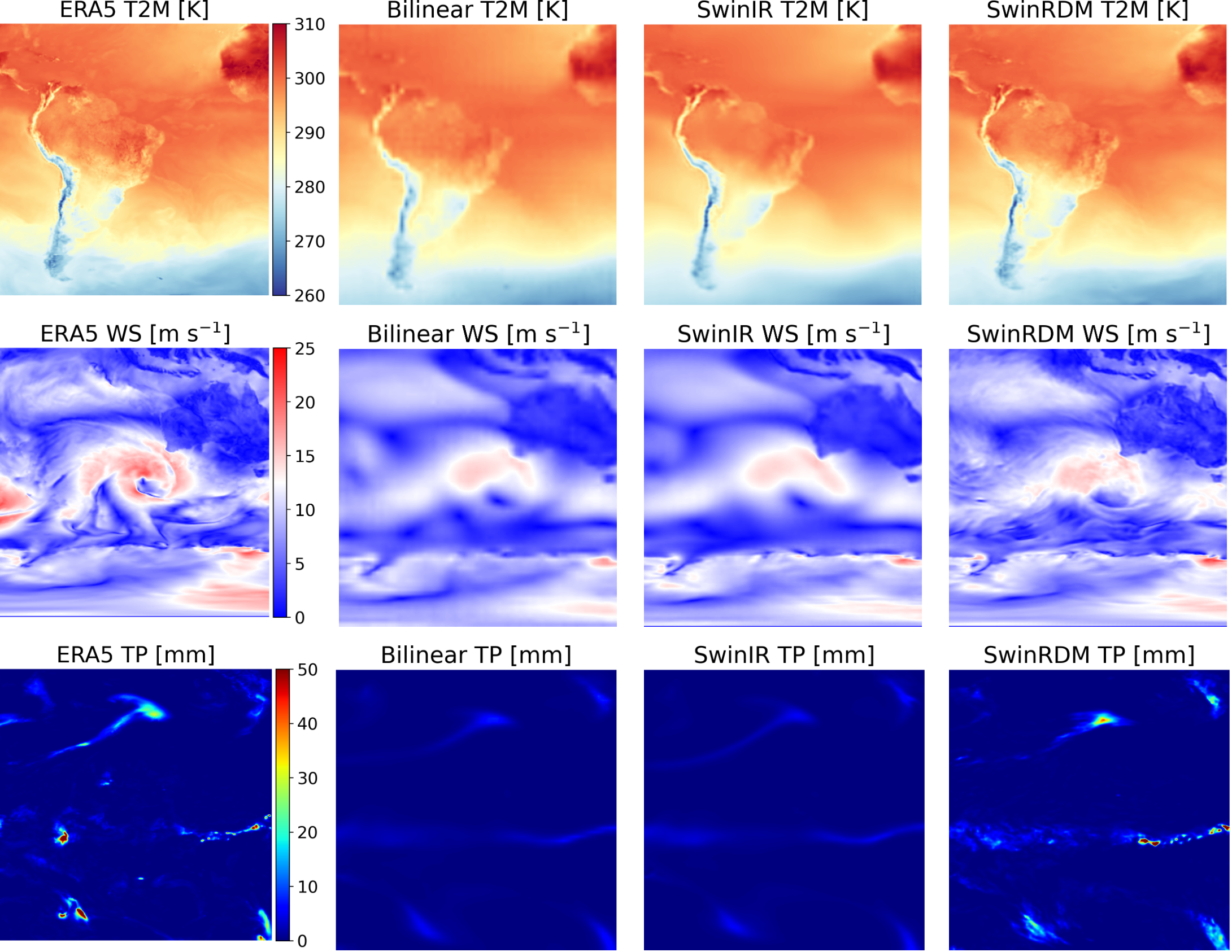}
    \caption{Visual comparison of example fields at the initialization time of June 4, 2018, 00:00 UTC. The first column shows the ERA5 fields for Z500, WS10, and TP at a lead time of 120 hours. The second to fourth columns represent the corresponding forecasting of different SR methods.}
    \label{fig:superres}
\end{figure}

\begin{table}[t]
 \centering
 \tabcolsep=5pt
 \begin{tabular}{l  c  c  c  c  c }
 \hline
  Methods & CSI2 & CSI5 & CSI10 & CSI20 & CSI50 \\
\hline
    Bilinear  &0.171 & 0.051 & 0.006 & 0.000 & 0.000 \\
    SwinIR  & 0.186 & 0.069 & 0.013 & 0.001 & 0.000  \\
    SwinRDM  & 0.246 & 0.179 & 0.111 & 0.046 & 0.010 \\
    SwinRDM*  & 0.262 & 0.190 & 0.123 & 0.049 & 0.006\\    
\hline
 \end{tabular}
 \caption{Critical success index (CSI) of six-hour accumulate total precipitation (TP) with different thresholds, i.e., 2 mm, 5 mm, 10 mm, 20 mm, and 50 mm.}
 \label{table:csi}
 \end{table}
 
 \begin{table}[t]
 \centering
 \tabcolsep=4pt
 \begin{tabular}{l  c  c  c  c  c c c c c c c}
 \hline
 Methods   & Z500 & T850 & T2M & TP  \\
\hline
    IFS & 154/334 & 1.36/2.03 & 1.35/1.77 & 2.36/2.59 \\
    SwinRNN & 207/392 & 1.39/2.05 & 1.18/1.63 & 2.01/2.14  \\
    SwinRNN+ & 152/316 & 1.12/1.75 & 0.99/1.42 & 1.88/2.07 \\
    SwinRDM & 156/316 & 1.23/1.83 & 1.07/1.49 & 2.02/2.24 \\
    SwinRDM* & \textbf{153/313} & \textbf{1.15/1.76} & \textbf{1.01/1.43} & \textbf{1.87/2.06} \\ 
\hline
 \end{tabular}
 \caption{Comparison with state-of-the-art IFS and SwinRNN. The RMSE scores for 3 and 5 days forecast times in the years 2017 and 2018 are shown. SwinRDM* is the 10-member ensemble version of SwinRDM. The units are the same as Table \ref{featfusion}.}
 \label{table:sota}
 \end{table}

\begin{figure}
  \centering
  \includegraphics[width=0.49\linewidth]{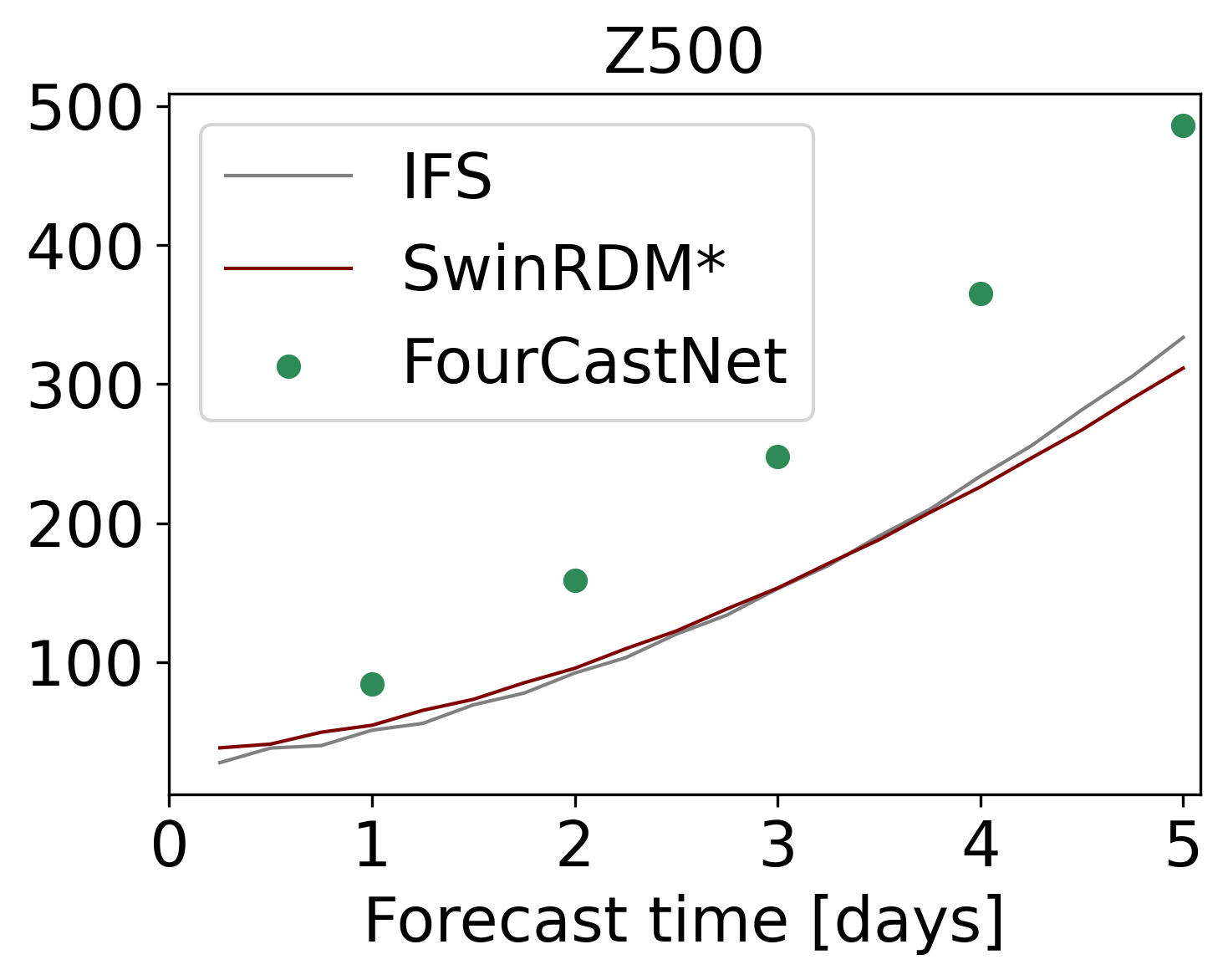} \hfill
  \includegraphics[width=0.49\linewidth]{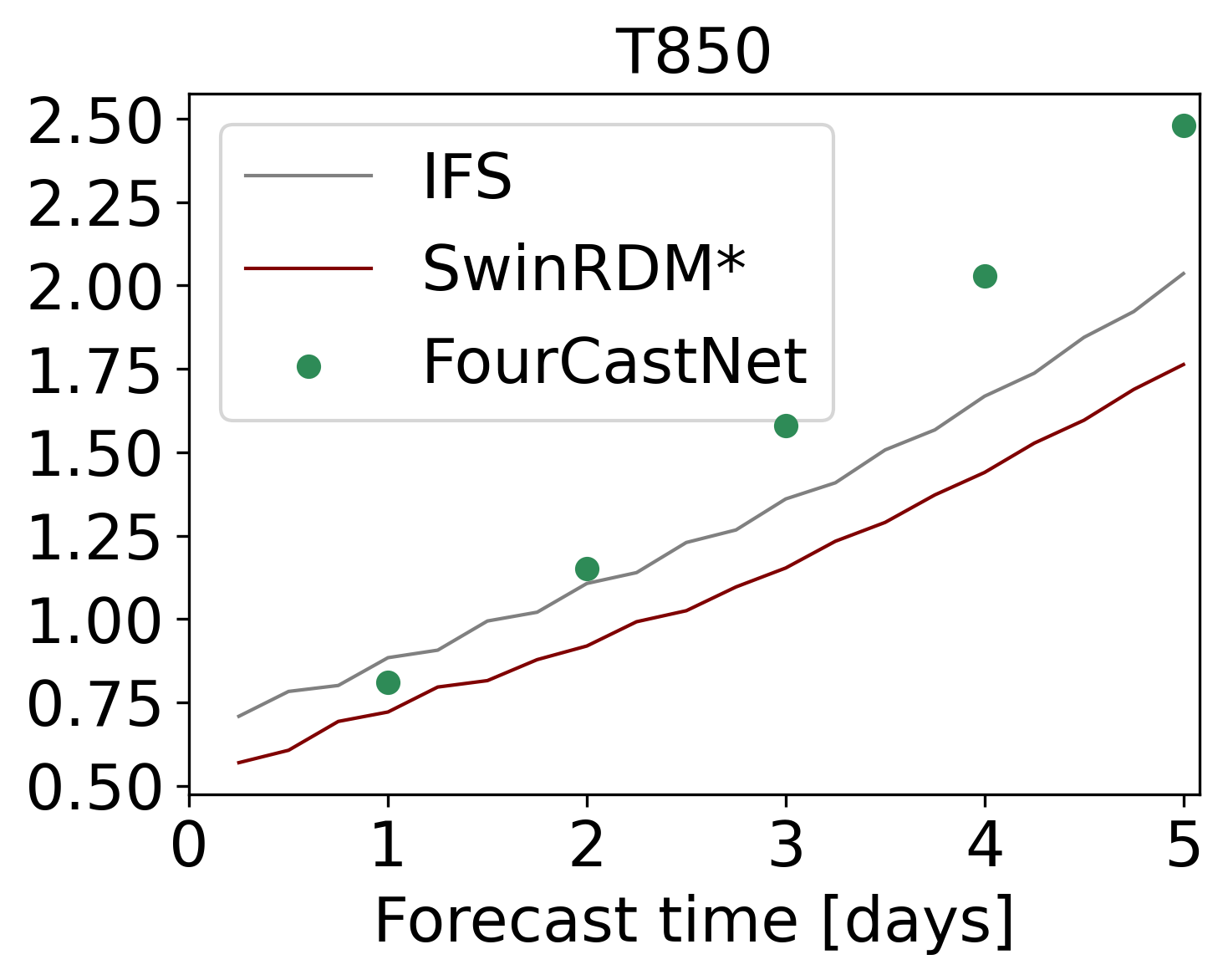}
  \caption{Comparison with IFS and FourCastNet on the test data of 2018. The RMSE of Z500 and T850 is shown.}
  \label{fig:sota}
\end{figure}

\begin{figure*}[t]
    \centering
    \includegraphics[width=\linewidth]{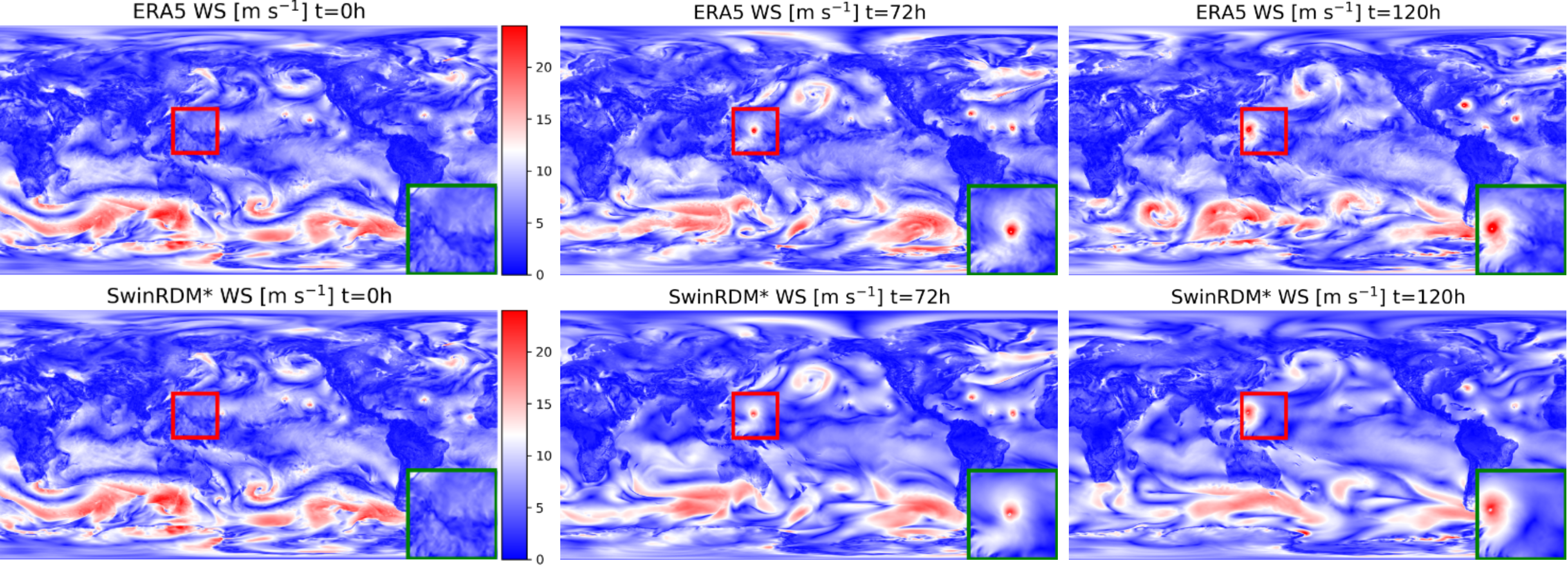}
    \caption{Qualitative illustration of a global near-surface wind forecast generated by our SwinRDM*. The prediction starts at the initial time of September 8, 2018, 06:00 UTC. The zoom-in area shows the beginning of Super Typhoon Mangkhut. Our method successfully forecasts Super Typhoon Mangkhut with high accuracy and rich fine-scale features.}
    \label{fig:qualitativews}
\end{figure*}

\subsection{Ablation Study on Diffusion-based SR}
In this subsection, we compare three variants of the super-resolution methods: bilinear interpolation, SwinIR \cite{liang2021swinir}, and our diffusion model. Bilinear interpolation simply upsamples the low-resolution output. SwinIR is a state-of-the-art SR method constructed by several residual Swin Transformer blocks. All methods are jointly trained with the same low-resolution forecasting model SwinRNN+. In addition to the RMSE metric, FID is used here to evaluate the ability to generate realistic results. FID is computed over six variables by modifying the input channels and weights of the Inception V3 model \cite{szegedy2016rethinking}.

\subsubsection{Diffusion-based SR achieves high visual quality}
As can be seen from Table \ref{table:superres}, there is little difference between different super-resolution methods in terms of the RMSE metrics of all variables. However, the FID metrics are significantly different. The bilinear interpolation obtains the worst FID score, and SwinIR slightly improves it by 8. Equipped with the diffusion-based super-resolution model, our SwinRDM considerably improves the FID score by 175, indicating that the diffusion-based super-resolution model can help generate high-quality and realistic results. SwinRDM* is a 10-member ensemble version of our SwinRDM, and it makes a good trade-off between the RMSE and FID. Figure \ref{fig:fid} shows the FID scores for different lead times. For a data-driven recurrent forecasting model, the predictions may become smoother with the increase in the lead time. The FID scores for bilinear interpolation and SwinIR increase with the forecast time, while our SwinRDM keeps low FID scores for lead times of up to 5 days. This again shows the good property of the diffusion-based super-resolution model. Although our SwinRDM* increases the FID scores of SwinRDM, it still maintains relatively stable FID scores compared with bilinear interpolation and SwinIR.


\subsubsection{High visual quality means high forecasting quality}
The qualitative results of different SR methods are shown in Figure \ref{fig:superres}. SwinRDM successfully captures small-scale structures and generates high-quality results at a 5-day lead time. By contrast, both Bilinear and SwinIR produce blurry results and fail to generate rich details. These details are essential for weather forecasting, especially for variables that have complex structures and variations, such as wind speed (WS) and TP. To further verify the superiority of our method, we calculate the critical success index (CSI) \cite{jolliffe2012forecast} for TP. The CSI can indicate the prediction performance under different thresholds. As shown in Table \ref{table:csi}, SwinRDM surpasses the strong baseline SwinIR by a large margin, which increases the CSI2, CSI5, and CSI10 by 6\%, 11\%, 10\%,  respectively. For higher CSI thresholds (20mm and 50mm), only our SwinRDM and SwinRDM* can get decent results, indicating that our methods can forecast extreme precipitation more effectively. Thus, our diffusion-based super-resolution model can generate high-resolution and high-quality forecasting results.

\subsection{Comparison with State-of-the-art Methods}
We compare our methods with state-of-the-art operational IFS and data-driven methods. Since the high-resolution IFS results are not available online due to the data center migration currently, we regrid our results to $5.625^{\circ}$ so that we can compare them to the IFS results provided by the WeatherBench \cite{rasp2020weatherbench}. Different methods can be directly compared due to the equivalence of evaluation on different resolutions. As stated in \cite{rasp2020weatherbench}, there is nearly no difference for evaluation at different resolutions. We have also verified this statement. For data-driven methods, we choose the SwinRNN \cite{hu2022swinvrnn} and FourCastNet \cite{pathak2022fourcastnet} for comparison. To the best of our knowledge, SwinRNN is the best method at $5.625^{\circ}$ resolution, and FourCastNet is the best method at $0.25^{\circ}$ resolution. Table \ref{table:sota} shows the comparison results for the years 2017 and 2018. Our SwinRNN+ method is trained on $1.40625^{\circ}$ resolution data, and it achieves significantly better performance compared to SwinRNN, showing the effectiveness of our improvement. The SwinRNN+ is also the first method that can surpass the IFS on both the surface-level and pressure-level variables. Our SwinRDM* trained for forecasting at $0.25^{\circ}$ resolution also shows better performance compared to the IFS at lead times of 3 days and 5 days. Specifically, at a lead time of 5 days, it outperforms the IFS by 21, 0.27, 0.34, and 0.53 in terms of Z500, T850, T2M, and TP, respectively. Since FourCastNet is only evaluated for the year 2018, we show the comparison results in terms of Z500 and T850 for the year 2018 in Figure \ref{fig:sota}. The results of FourCastNet are obtained from the original paper. As shown in the figure, our SwinRDM* method shows better performance compared with FourCastNet, indicating that our method shows state-of-the-art performance at $0.25^{\circ}$ resolution. Note that our SwinRDM* shows slightly lower performance at lead times less than 3 days compared to IFS. This may attribute to the lower representational power of the encoder since we find that the capacity of the encoder is important to the short-range forecasting performance. How to better extract the historical context information is essential for the encoder, and we leave this for future research.


\subsection{Qualitative Illustration}
Figure \ref{fig:qualitativews} shows qualitative results of the proposed SwinRDM*. Our model is tested on 8 September 2018, 06:00 UTC to forecast the near-surface wind speeds (WS) at lead times of 3 days and 5 days. The wind speeds are computed from the predicted zonal and meridional components of the wind velocity i.e., $WS=\sqrt{U_{10}^2 + V{10}^2}$. The results of ERA5 are the ground truth. As shown in the figure, our method can forecast wind speeds for up to 5 days with high resolution and high quality. Specifically, we can see from the zoom-in area in the figure that our method can successfully forecast and track Super Typhoon Mangkhut. The ability to forecast this kind of extreme event is really beneficial for the mitigation of loss of life and property damage. Our method shows high forecasting accuracy and the ability to capture fine-scale dynamics at high resolution.

\section{Conclusion}
We propose a high-resolution data-driven medium-range weather forecasting model named SwinRDM by integrating SwinRNN+ with a diffusion-based super-resolution model. Our SwinRNN+ is improved upon SwinRNN by trading the multi-scale design for higher feature dimensions and adding a feature aggregation layer, which achieves superior performance compared to the operational NWP model on all key variables at $1.40625^{\circ}$ resolution and lead times of up to 5 days. Our SwinRDM uses a diffusion-based super-resolution model conditioned on the forecasting results of SwinRNN+ to achieve high-resolution forecasting. The diffusion model helps generate high-resolution and high-quality forecasting results and can also perform ensemble forecasting. The experimental results show that our method achieves SOTA performance at $0.25^{\circ}$ resolution.

\bibliography{aaai23}

\end{document}